\documentclass{article} 
\usepackage{hyperref}
\usepackage{ucs}
\usepackage{comment}
\usepackage{amsmath}
\usepackage{bbold}
\usepackage{pbox}
\usepackage{tcolorbox}
\usepackage[font=small]{caption}
\usepackage{capt-of}
\usepackage{bm}
\usepackage{graphicx}
\usepackage{cancel}
\usepackage{array}
\usepackage{multirow}
\usepackage{colortbl}
\usepackage{tikz}
\usepackage{xspace}
\usepackage{subfig}
\usepackage{algorithm}
\usepackage[noend]{algpseudocode}
\usepackage{soul}
\usepackage{enumitem}
\usepackage{comment}

\newcommand{\mdp}{{\sc mdp}\xspace}

\newcommand{\gnn}{{\sc gnn}\xspace}

\newcommand{\gnns}{{\sc gnn}s\xspace}

\newcommand{\sac}{{\sc sac}\xspace}

\newcommand{\mlp}{{\sc mlp}\xspace}

\newcommand{\sr}{{\sc SR}\xspace}
\newcommand{\rlearning}{RL\xspace}

\newcommand{\sgn}{{\sc s-gn}\xspace}
\newcommand{\sint}{{\sc s-in}\xspace}
\newcommand{\srn}{{\sc s-rn}\xspace}
\newcommand{\sds}{{\sc s-ds}\xspace}
\newcommand{\sflat}{{\sc s-flat}\xspace}
\newcommand{\cgn}{{\sc c-gn}\xspace}
\newcommand{\cint}{{\sc c-in}\xspace}
\newcommand{\crn}{{\sc c-rn}\xspace}
\newcommand{\cds}{{\sc c-ds}\xspace}
\newcommand{\cflat}{{\sc c-flat}\xspace}
\newcommand{\gn}{{\sc gn}\xspace}
\newcommand{\inter}{{\sc in}\xspace}
\newcommand{\rnet}{{\sc rn}\xspace}
\newcommand{\ds}{{\sc ds}\xspace}

\newcommand{\ai}{{\sc ai}\xspace}
\newcommand{\acl}{{\sc acl}\xspace}

\newcommand{\her}{{\sc her}\xspace}
\newcommand{\mcher}{{\sc mc-her}\xspace}

\newcommand{\lp}{{\sc lp}\xspace}

\newcommand\wi[1]{$\circ$}
\newcommand\bu[1]{$\bullet$}
\newcommand\ot[1]{$\star$}
\newcommand\spa[1]{$\spadesuit$}
\newcommand\dn[1]{.}
\newcommand\no[1]{}
\newcommand\bo[1]{$\bullet\star$}

\usepackage{array}

\newcounter{cbox} 
\newcommand{\cbox}{\arabic{cbox}}

\newcounter{cmes} 
\newcommand{\cmes}{\arabic{cmes}}

\usepackage{xcolor}
\definecolor{myred}{rgb}{0.8,0,0}
\definecolor{mypurple}{rgb}{0.6,0.22,0.8}

\definecolor{mygreen}{rgb}{0,0.6,0}
\definecolor{myblue}{rgb}{0,0,0.7}
\definecolor{myindigo}{rgb}{0.4,0,0.7}

\usepackage{collas2022_conference,times}

\usepackage{amsmath,amsfonts,bm}

\def\eqref#1{equation~\ref{#1}}

\def\1{\bm{1}}

\DeclareMathAlphabet{\mathsfit}{\encodingdefault}{\sfdefault}{m}{sl}
\SetMathAlphabet{\mathsfit}{bold}{\encodingdefault}{\sfdefault}{bx}{n}

\usepackage{hyperref}
\usepackage{wrapfig}
\usepackage{url}

\title{Learning Object-Centered Autotelic Behaviors \\with Graph Neural Networks}

\author{Ahmed Akakzia \\
Sorbonne Université\\
\texttt{ahmed.akakzia@isir.upmc.fr} \\
\AND
Olivier Sigaud \\
Sorbonne Université \\
}

\collasfinalcopy 
\begin{document}

\maketitle

\begin{abstract}
Although humans live in an open-ended world and endlessly face new challenges, they do not have to learn from scratch each time they face the next one. Rather, they have access to a handful of previously learned skills, which they rapidly adapt to new situations. In artificial intelligence, autotelic agents\,---\,which are intrinsically motivated to represent and set their own goals\,---\,exhibit promising skill adaptation capabilities. However, these capabilities are highly constrained by their policy and goal space representations. In this paper, we propose to investigate the impact of these representations on the learning and transfer capabilities of autotelic agents. We study different implementations of autotelic agents using four types of Graph Neural Networks policy representations and two types of goal spaces, either geometric or predicate-based. By testing agents on unseen goals, we show that combining object-centered architectures that are expressive enough with semantic relational goals helps learning to reach more difficult goals. We also release our graph-based implementations to encourage further research in this direction.
\end{abstract}

\section{Introduction}
A central challenge in artificial intelligence (\ai) consists in designing artificial agents capable of solving an unrestricted set of tasks in a continual and open-ended skill learning process. In principle, these processes should be domain-agnostic. Reinforcement learning (\rlearning) seems to be an adequate paradigm to solve a single sequential decision problem from a reward signal \citep{sutton1999reinforcement}.  Nevertheless, this signal is usually predetermined and highly grounded to its designer’s aspirations. Thus, the extension of the \rlearning framework to an open-ended sequences of unpredictable tasks raises difficult questions.

Recently, a promising line of research has been interested in the design of \textit{autotelic agents}, borrowing older ideas from \citep{steels2004autotelic}. These agents are intrinsically motivated to represent, set and pursue their own goals. Usually, they do not depend on any external reinforcement signal, since they autonomously reward themselves over the completion of their own goals. Autotelic agents are known to be open-ended learners. Through \rlearning, they manage to acquire goal-directed behaviors which can transfer to domains sharing similar goal spaces. However, this transfer is deeply bound to their representational capabilities. 

From that perspective, a key challenge consists in endowing autotelic agents with appropriate inductive biases to enhance their representational power. To enable efficient transfer, such biases should express a set of general and structured features. On the one hand, the design of the autotelic agents’ goal spaces should 
leverage the power of structured semantic representations.
Namely, recent works in AI \citep{akakzia2021grounding, alomari2017natural, kulick2013active, tellex2011approaching} introduced symbolic high-level object-centered representations to explicitly capture abstract spatial relations such as proximity and aboveness, where the latter is used to refer to the quality of being directly above. By contrast,  other works use plain spatial target coordinates specific to each of the available objects \citep{colas2019curious, li2019towards, lanier2019curiosity}.

On the other hand, although neural networks are flexible tools to learn latent representations, their raw usage is insufficient to capture disentangled representations from high-dimensional structured input. Recently, Graph Neural Networks (\gnns) have been introduced to implement relational inductive biases in neural networks. They mainly rely on shared networks to transfer features among the input components. Besides, they follow efficient computation schemes: through their neighborhood aggregation and graph-level pooling schemes, they easily capture the existing relationships between nodes. 

\paragraph{Contributions. }In this paper, we provide a systematic study of the use of \gnns in autotelic learning within a multi-object manipulation domain. More specifically, we investigate 4 variants of \gnns: \textit{full graph networks}, \textit{interaction networks}, \textit{relation networks} and \textit{deep sets}. Furthermore, we consider two different types of goal spaces: 1) \textit{semantic goals} based on binary predicates describing spatial relations between physical objects; 2) \textit{continuous goals} corresponding to specific target positions for each object. Finally, we assess the transfer capabilities of the best performing \gnn-based agents by introducing three sets of held-out semantic goals defining three different scenarios: 1) transfer to combinations of configurations (such as a stack and a pyramid); 2) transfer from goals based on pair-wise relations to goals based on triple-wise relations (pyramids); 3) transfer to higher order stacking of objects.

Our results show that 
\begin{itemize}
    \item Semantic goal spaces induce a higher level of abstraction than continuous goal spaces, enabling lighter \gnn-based architecture to perform on par with the ones that use the whole computational scheme.
    \item Performing the edge update step is sufficient for good transfer to combinations of previously seen goals.
    \item Node updates promote the transfer from goals based on pair-wise relations to goals based on triple-wise relations, as information flow not only between pairs, but also between all the nodes.
    \item Relation networks outperform full graph and interaction networks in transferring to higher order stacks on objects.
\end{itemize}

These results suggest that coupling semantic goal spaces with sufficiently representative graph-based networks helps to learn more complex goals and yields better transfer capabilities. Finally, we release our implementations of the considered \gnn-based architectures in multi-object manipulation domain to encourage further research in this direction\footnote{\href{https://github.com/akakzia/rlgraph_2}{https://github.com/akakzia/rlgraph\_2}.}


\section{Related work}

This paper relies on several previous works from different areas of research within \ai. Namely, we consider recent findings in automatic curriculum learning, semantic goal representations, graph neural networks and graph-based autotelic learning, and combinations of several of these aspects.

\paragraph{Automatic Curriculum Learning. }Adaptability is a key characteristic enabling humans to display an exceptional capacity to learn \citep{elman1993learning} and works in \ai attempted to leverage similar automatic curriculum learning (\acl) schemes in artificial agents \citep{portelas2020automatic}. Most of these approaches leverage forms of intrinsic motivations to power their exploration and learning progress (\lp) \citep{bellemare2016unifying, achiam2017surprise, nair2018visual, burda2018large, pathak2019self, colas2019curious, pong2019skew}. In this paper, our agents borrow the \lp-based curriculum learning algorithm introduced in \cite{colas2019curious} when targeting continuous goals, but we show this in not necessary when targeting semantic goals.

\paragraph{Semantic Goal Representations. }
Studies in developmental psychology suggest that notions such as proximity, animacy and containment are innately grounded in the perceptual world of the infant \citep{mandler2012spatial}. Inspired by this line of thought, recent works in \ai introduced symbolic high-level representations to explicitly capture abstract spatial relations \citep{tellex2011approaching, kulick2013active, alomari2017natural,akakzia2021grounding}. We borrow the semantic goal representations used in \cite{akakzia2021grounding} and based on the predicates \textit{close} and \textit{above}. Such semantic representations are more abstract the classic goal-as-state representations, as they account for the underlying relations between objects independently of their perceived states, such as their geometric positions.

\paragraph{Graph Neural Networks. } \gnns are powerful tools to implement strong inductive biases that focus on structured representations \citep{battaglia2018relational}. At the price of more computations, they efficiently foster combinatorial generalization and improve sample efficiency over standard architectures in different machine learning domains \citep{gilmer2017neural, scarselli2005graph, zaheer2017deep, li2019towards}. \gnns parse the stream of input features into several objects, called \textit{nodes}. They also capture the relational features between pairs of these objects which they store in the corresponding \textit{edges}. They usually involve three computational schemes: 1) \textbf{Edge updates} using the initial features of the edge and both features of the nodes involved within that edge; 2) \textbf{Node updates} using the initial features of the node and the aggregated features of the edges that enter that nodes; 3) \textbf{Graph output} using an aggregation of either all the nodes or the edges features. The first two steps involve shared networks, which enable transfer between the different nodes and edges. Depending on the order and the nature of the computational steps, there exist many variants of \gnns. In this paper, we only consider 4 of these variants: \textit{full graph networks} \citep{battaglia2018relational}, \textit{interaction networks} \citep{battaglia2016interaction}, \textit{relation networks} \citep{santoro2017simple} and \textit{deep sets} \citep{zaheer2017deep}. In general, these variants are shown to outperform flat architectures when combined with object-centered representations. Details about the implementations of these variants are provided in Section~\ref{sec:graph_autotelic}.

\paragraph{Graph-based Autotelic \rlearning. } \gnns have been used to solve \rlearning problems \citep{zambaldi2018relational, li2019towards, colas2020language, akakzia2021grounding}. By contrast to \cite{li2019towards, colas2020language, akakzia2021grounding}\,---\,which explicitly associate a node to each object in an object manipulation domain\,---\,the approach in \cite{zambaldi2018relational} attempts to solve the StarCraft II mini-games \citep{vinyals2017starcraft} without object-centered inductive bias. In the latter, the nodes do not correspond to specific objects, but rather to randomly scattered boxes of pixels. In this paper, we rather join the former group.

\paragraph{Structured Policies and Representations.} Close to our work, \citep{bapst2019structured} study the combination of structured representations and graph-based policies and show that it outperforms setups that use less induced structure. On the one hand, like us, they consider both continuous and semantic settings. In the former, while they add a node for each target goal, we encode continuous goal features within the edges of our graphs. In the latter, while they use an additional conversion layer to feed the policy with the converted geometric features, we directly use binary semantic predicates as inputs to both our critics and actors. On the other hand, by contrast to their graph computation scheme which involves an encode-process-decode architecture \citep{battaglia2016interaction}, our graphs are simpler as they do not use any form of recurrence. In fact, through only one step of computation involving one edge update and one node update, inputs are converted into either actions (for the actor) or q-values (for the critic). Finally, in this paper, we consider many types of \gnns which use different computation schemes and we aim at assessing their transfer capabilities to previously held-out goals. 

\section{Methods}

In this section, we state the problem we address in this paper, then we introduce the object manipulation environment and the two goal spaces that we consider (Section~\ref{sec:env_goals}). Finally, we present the graph-based implementations of our autotelic agents (Section~\ref{sec:graph_autotelic})

\subsection{Problem statement}
\label{sec:pb_statement}

We address hard exploration problems where an agent is expected to learn a large diversity of complex behaviors. We cast the problem into the framework of goal-conditioned reinforcement learning  \citep{colas2020intrinsically} and particularly consider autotelic agents which can represent, set and pursue their own goals. These agents learn from a sparse reward signal, i.e. they are only rewarded for reaching the goal they have set.
More formally, these sparse reward autotelic agents are facing a rewardless Markov Decision Process \mdp$= \{S, A, T\}$ where states $s \in S$ and actions $a \in A$ are continuous valued and the transition function $T: S\times A \rightarrow \Pi(S)$ defines the probability of reaching any state after performing an action from a state. Agents themselves are implemented as a goal-conditioned policy $\pi(a|s,g)$ and are rewarded with a function $r(s,g)$ which determines whether state $s$ satisfies goal $g$. The key feature of autotelic agents is that they choose on which goal $g$ to work at any moment. In this paper, goals are either semantic, i.e. they are represented as a set of binary predicates describing the features of the scene that matter for the agent's tasks, or continuous, i.e they directly correspond to a subset of the features perceived by the agent.

\subsection{Environment and goal spaces}
\label{sec:env_goals}

\paragraph{The Fetch Manipulate Environment. }All agents studied in this paper evolve in the \textit{Fetch Manipulate} domain from  \citet{akakzia2021grounding}, which is a variant of the standard \textit{Fetch} domains \citep{plappert2018multi}. We extend it to a 5-object setup: the agent is a 4-DoF robotic arm facing 5 colored objects on a table. It perceives features of its body and of the surrounding objects. These features include geometric positions, orientations and velocities.

\paragraph{Semantic versus continuous Goals. } From the perceived features, agents using semantic goals build high-level binary representations that assert the presence (1) or absence (0) of the binary spatial relations \textit{above} and \textit{close} between objects. As the latter is symmetric (\textit{close(A, B) = close(B, A)}), we only consider 10 combinations of objects for this predicate. However, we consider all the 20 ordered pairs of objects for the \textit{above} predicate. This yields semantic goal vectors of 30 dimensions. The resulting configuration space contains $2^{30}$ elements, among which $\sim 75.000$ are physically reachable. These semantic representations are inspired by the work of \cite{mandler2012spatial} on a minimal set of spatial primitives children seem to be born with, or to develop early in life. Initially empty, the set of discovered semantic goals gets gradually filled each time an agent encounters new configurations.

\begin{wrapfigure}[6]{r}{0.35\textwidth}
     \vspace{-0.6cm}
     \centering
    \includegraphics[width=0.15\textwidth]{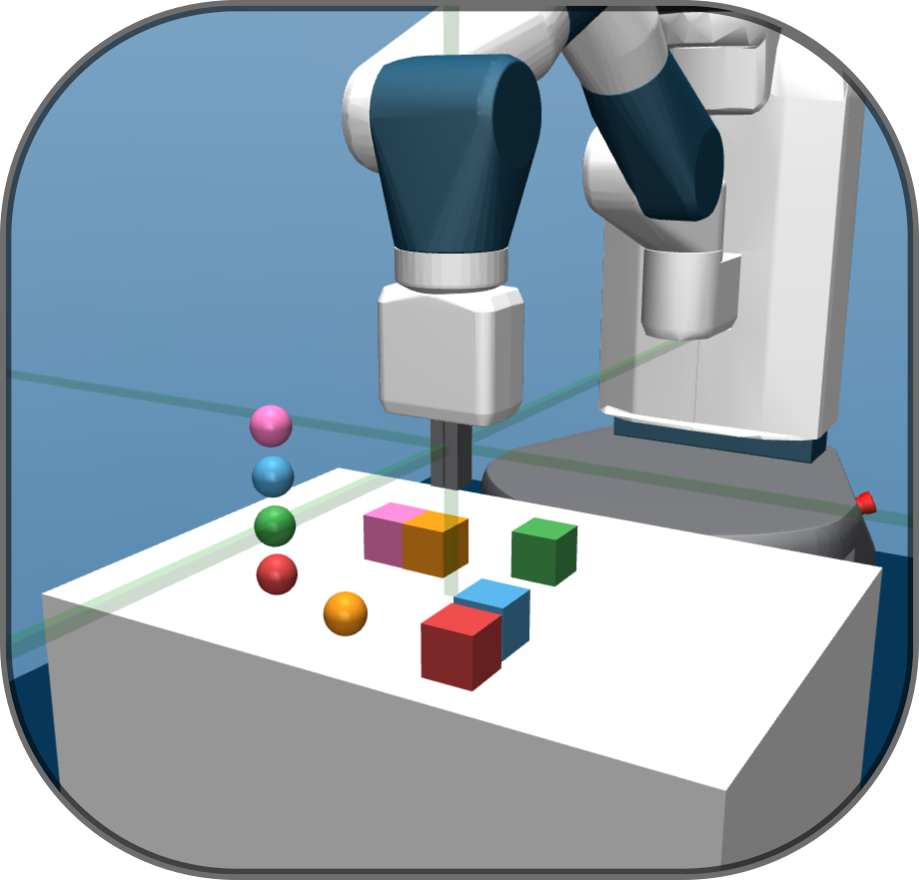}
    \caption{Illustration of objects and targets.}
    \label{fig:fetch_targets}
\end{wrapfigure}
By contrast with semantic goals, continuous goals directly use the perceived features, i.e. goals correspond to precise target positions for each available object. To succeed, agents have to place every object in its corresponding target position, see Figure~\ref{fig:fetch_targets} for an illustration. These goal spaces are used in many works attempting to solve multi-object manipulation problems \citep{colas2019curious, li2019towards, lanier2019curiosity}.

\vspace{0.6cm}
\paragraph{Autotelic learning. } All studied agents autonomously select and attempt to master goals from the set of discovered goals. Agents using semantic goals simply reward themselves for each object for which all the predicates involving that object are verified. An episode ends successfully if all predicates about all objects are verified before a time limit. At the beginning of an episode, the blocks are procedurally placed on the table so that they are never initially stacked. 


By contrast, the autotelic learning process of agents using continuous goals is more involved. First, we assume that these agents are initially aware that they can construct stacks using the available objects, and that the maximum number of objects stacked corresponds to the number of available objects. Second, at the beginning of each episode, these agents autonomously select how many objects they want to stack (from 0 up to 5 in this paper). Accordingly, target positions are generated for each object. These agents reward themselves for each object placed correctly within a range of its corresponding target position. An episode ends successfully if all objects are placed correctly before a time limit. To further accelerate the learning process, we consider biased initializations as part of a way to adapt the difficulty of the task to the learner's skills: at the beginning of each episode, and with a probability of 0.2, blocks are arranged into a stack of up to 5 objects. 
We call this non-trivial scene reset.
To stabilize the learning process, we use an automatic \lp-based curriculum \citep{colas2019curious}: based on their learning progress estimations, agents can choose to target goals with no stacks, a stack of 2, 3, 4 or 5 objects where all target positions that are not involved in stacks are automatically generated directly on the table. As shown in Appendix~\ref{supp:add_curriculum}, calling upon this additional \acl process is necessary for learning to work when using continuous goals.

\subsection{Graph-based autotelic learning}
\label{sec:graph_autotelic}
In this section, we describe the implementation of the intrinsically motivated goal-conditioned \rlearning module using \gnns. This module is powered by the Soft-Actor Critic algorithm (\sac) \citep{haarnoja2018soft} where both the critic and the policy networks are \gnns. We use the Multi-criteria Hindsight Experience Replay algorithm (\mcher) to facilitate transfer between goals \citep{lanier2019curiosity}. \mcher extends the Hindsight Experience Replay (\her) \citep{andrychowicz2017hindsight} strategy to multi-object scenarios, enabling further transfer between partial features of the goal vector.

\subsubsection{Graph structure}
\label{sec:graph_structure}
\begin{figure}[!hbt]
  \centering
  \subfloat[Input edge features as target predicates where $p_1=$close and $p_2=$above.\label{fig:graph_sem}]{\includegraphics[width=0.4\linewidth]{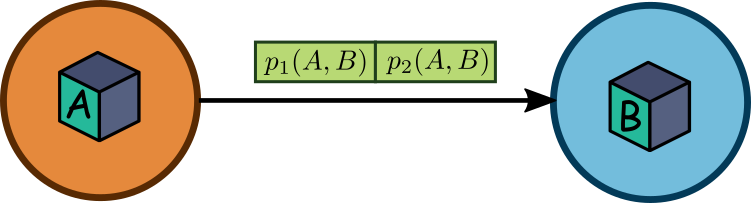}}\hfill
  \subfloat[Input edge features as target geometric positions.\label{fig:graph_con}]{\includegraphics[width=0.395\linewidth]{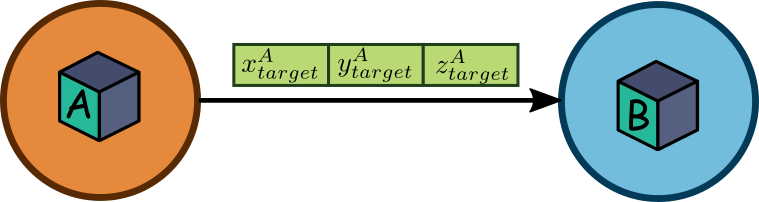}}
  \caption{Illustration of a single directed edge for semantic goals (a) and continuous goals (b).}
  \label{fig:graph_edge}
\end{figure}
All our agents use a fully connected graph structure: every object corresponds to a node, and all nodes are connected. First, each node holds the features of a particular object in the scene. Second, each edge linking a source and a recipient node holds partial features of the goal. As illustrated in Figure~\ref{fig:graph_edge},  for semantic goals, these features correspond to the predicates that involve both the source and the recipient node, while for continuous goals, they correspond to the target position of the block corresponding to the source node. Finally, the global features correspond either to the agent's body state (in the case of the policy) or to a concatenation of the agent's body state and the action (in the case of the critic). We respectively denote the node features, edges features and global features with $X$, $E$ and $U$.

\subsubsection{Graph computations}
\label{sec:graph_computations}
\begin{figure}[!ht]
 \centering
    \includegraphics[width=\textwidth]{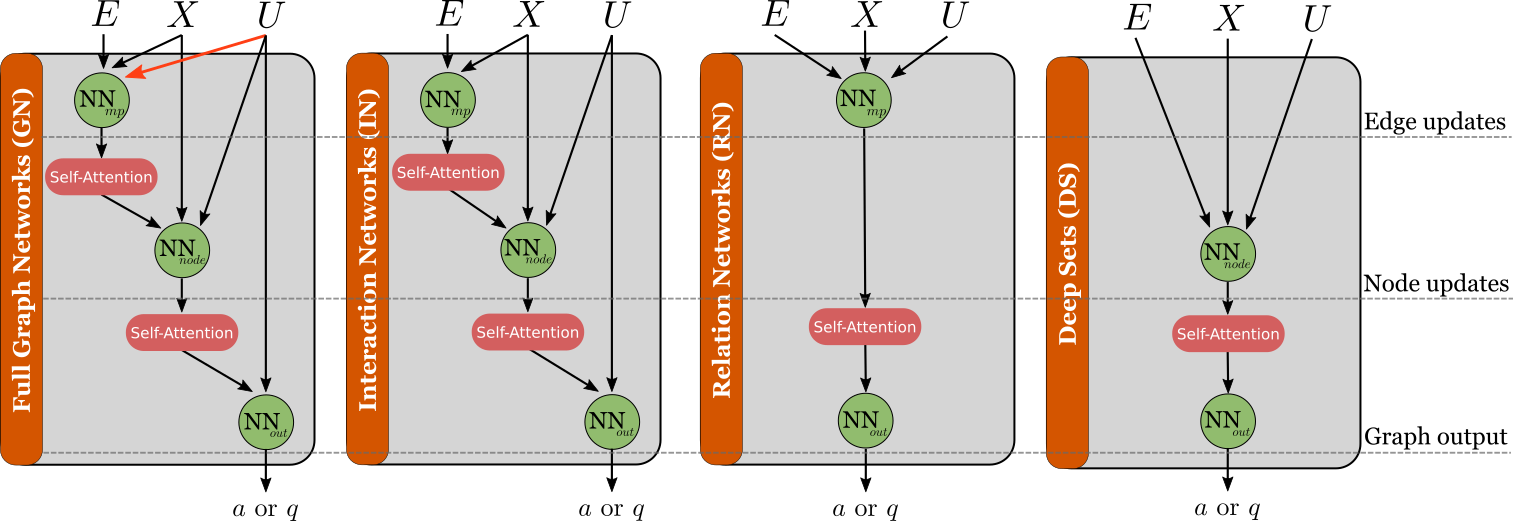}
    \caption{Illustration of the different computational schemes for (from left to right) \gn, \inter, \rnet and \ds. $E$, $X$ and $U$ respectively correspond to the edge features, node features and global features. Note that \gn uses $U$ to update edges features (red arrow), while \inter does not and \rnet only updates edges features, while \ds only updates nodes features.}
    \label{fig:graphs}
\end{figure}

Although all our agents rely on the same graph structure, they use different computation schemes. In this paper, we focus on four particular types of \gnns: full graph networks (\gn), interaction networks (\inter), relation networks (\rnet) and deep sets (\ds). Figure~\ref{fig:graphs} illustrates the different computation steps for each architecture.

\paragraph{Full Graph Network (\gn). }As its name suggests, this architecture uses the whole computation scheme within a standard graph network block. See Figure~\ref{fig:graphs} for an illustration. First, an \textit{edge update} step is performed. A shared network $NN_{mp}$ is used to compute the update features of each edge. It takes as input the concatenated input features of each edge (goal features), the involved source and recipient nodes (object features) and the global features. Second, a \textit{node update} step is performed for each node using a second shared network $NN_{node}$. It takes as input the concatenated input features of the considered node, the global features and an aggregation of the updated features of the incoming edges. Third, the \textit{graph output} step is performed, where the updated features of the nodes are pooled, concatenated with the global features and fed to a readout network $NN_{out}$. The output quantity corresponds to either the action (in the case of the actor) or the q-value (in the case of the critic). In this paper, we use self-attention to compute the weighing scores used in all the aggregation steps \citep{vaswani2017attention, velivckovic2017graph}.  

\paragraph{Interaction Network (\inter). }This architecture resembles the one described in the \gn architecture. The only difference is that, during the edge update step, the global features are not used as inputs to the shared network $NN_{mp}$.

\paragraph{Relation Network (\rnet). }This architecture entirely bypasses the node update step. It only performs the edge update step using the shared network $NN_{mp}$, which takes as inputs the initial node, edge and global features. The output vector is aggregated using a self-attention module, then fed to a readout network $NN_{out}$.

\paragraph{Deep Sets (\ds). }This architecture entirely bypasses the edge update step. It only performs node updates using the shared network $NN_{node}$. The latter takes as input the node, edge and global features, outputs a vector is later fed to a self-attention module to compute attention scores. Finally, the aggregated vector is fed to a readout network $NN_{out}$.

\subsubsection{Pseudo code}

The autotelic learning mechanisms with respectively semantic and continuous goals are presented in Algorithms~\ref{algo:semantic} and \ref{algo:continuous}. Algorithms~\ref{algo:semantic_sample} and \ref{algo:update} further describe how goals are sampled and updated in Algorithm~\ref{algo:semantic}.

\begin{minipage}{0.49\textwidth}
\begin{algorithm}[H]
    \caption{~ Learning Semantic Goals}
    \label{algo:semantic}
	\begin{algorithmic}[1]
	\State \textbf{Require} Env $E$, number of trajectories per step $n$, replay function $R_{mcher}$
	\State Initialize policy $\Pi$, Uniform goal sampler $\mathcal{G}^s_{unif}$, buffer $B$.
	\State $L_{discovered}$ = []
	\Loop 
        \State $L_{goals} \leftarrow$ {\sc Sample goals($L_{discovered}$, $n$)}
        \State $L_{trajectories} \leftarrow E$.rollout($\Pi$,  $g$)
        \State $L_{discovered} \leftarrow $ {\sc Update goals}($L_{trajectories}$)
        \State $B$.update($L_{trajectories}$)
        \State $\mathcal{D}_{train} \leftarrow B$.sample($R_{mcher})$
        \State $\Pi$.update($\mathcal{D}_{train}$)
    \EndLoop
    \State\Return $\Pi$
	\end{algorithmic}
\end{algorithm}
\end{minipage}
\begin{minipage}{0.49\textwidth}
\begin{algorithm}[H]
    \caption{~ Learning Continuous Goals}
    \label{algo:continuous}
	\begin{algorithmic}[1]
	\State \textbf{Require} Env $E$, Goal classes $C_g$, number of trajectories per step $n$, replay function $R_{mcher}$
	\State Initialize policy $\Pi$, LP-based goal sampler $\mathcal{G}^s_{LP}$, buffer $B$.
	\Loop 
	   \State $L_{classes} \leftarrow \mathcal{G}^s_{LP}$.sample\_classes($C_g$, $n$)
	   \State $L_{goals} \leftarrow generate\_positions(L_{classes})$
        \State $L_{trajectories} \leftarrow E$.rollout($g$)
        \State $\mathcal{G}^s_{LP}$.update($L_{trajectories}$)
        \State $B$.update($L_{trajectories}$)
        \State $\mathcal{D}_{train} \leftarrow B$.sample($R_{mcher})$
        \State $\Pi$.update($\mathcal{D}_{train}$)
	\EndLoop
    \State\Return $\Pi$
	\end{algorithmic}
\end{algorithm}
\end{minipage}

\begin{minipage}{0.49\textwidth}
\begin{algorithm}[H]
    \caption{~{\sc Sample goals}}
    \label{algo:semantic_sample}
	\begin{algorithmic}[1]
	\State \textbf{Require} discovered goals list $L_{discovered}$, number of goals to sample $n$,
	\State $L_{samples}$ = []
	\For{$i$ in [1, .., $n$]}
	\If{$L_{discovered}$ is empty}
	\State $L_{samples}$.append(zeros)
	\Else 
	\State $g \leftarrow L_{discovered}$.sample\_uniform()
	\EndIf
	\EndFor
    \State\Return $L_{samples}$
	\end{algorithmic}
\end{algorithm}
\end{minipage}
\begin{minipage}{0.49\textwidth}
\begin{algorithm}[H]
    \caption{~ {\sc Update goals}}
    \label{algo:update}
	\begin{algorithmic}[1]
	\State \textbf{Require} Trajectory list $L_{\tau}$, discovered goals list $L_{discovered}$ $\mathcal{G}^s_{LP}$, buffer $B$.
	\For{$\tau$ in $L_{\tau}$}
	\State $(s, a, s', g_{achieved}, g_{desired}) \leftarrow \tau$
    \State Last\_$g_{achieved} \leftarrow g_{achieved}[-1]$
    \If{Last\_$g_{achieved}$ not in $L_{discovered}$}
    \State $L_{discovered}$.append(Last\_$g_{achieved}$)
    \EndIf
    \EndFor
    \State\Return $L_{discovered}$
    \State
	\end{algorithmic}
\end{algorithm}
\end{minipage}

\section{Experiments and results}
We first describe the experimental setup used in this paper. Then, we present the results obtained when training autotelic agents with semantic and continuous goals. Finally, we assess the transfer capabilities on the different architectures with three different scenarios. Additional studies and ablations are provided in the appendices.

\subsection{Experimental setup}
\label{sec:setup}

We train 4 graph-based autotelic agents in the \textit{Fetch Manipulate} domain with 5 objects using the graph architectures described in Section~\ref{sec:graph_autotelic}. We consider both the semantic and continuous goal spaces introduced in Section~\ref{sec:env_goals}.

\paragraph{Evaluation Classes. }To evaluate the agents, we define several evaluation classes for both semantic and continuous goals. First, for semantic goals, we consider classes of configurations where exactly $i$ pairs of blocks are close ($C_i$), configurations containing stacks of size $i$ ($S_i$), configurations containing pyramids of size 3 ($P_3$) and combinations of these. These classes are disjoint and their union does not cover the entire semantic configuration space, but they are representative enough and they enable fair comparisons between the agents. Second, for continuous goals, we consider classes of configurations where there are no stacks and where there is a stack of size $i$ ($\widetilde{S}_i$, where the symbol $\sim$ is for continuous).

\paragraph{Evaluation Metrics. }Evaluations are performed each 50 cycles. During one cycle, the agents perform 2 rollouts of 200 timesteps with 2 goals sampled autonomously. At test time, the per-class performance of the agent is computed on 24 goals of each evaluation class (264 semantic goals and 120 continuous goals). The measure of the agent's global success rate (\sr) is the average of all the per-class successes. Testing is conducted offline and with deterministic policies. 

\paragraph{Baseline. }For both semantic and continuous goals, we consider a flat baseline, where all the perceived features are concatenated and directly fed to the neural networks. We call Semantic-Flat (\sflat) and Continuous-Flat (\cflat) the flat baseline using respectively semantic and continuous goals. 

\paragraph{Networks Capacity. } Independently of their computational scheme, we make sure all the agents have the same networks capacity in terms of number of parameters to be optimized. As full graph networks use the highest number of parameters in principle, we provide the other agents with sufficient budget to match them. Concretely, we add an additional node updater and edge updater for respectively deep sets and relation networks architecture, and we make the flat networks sufficiently deep.

\begin{wraptable}[9]{r}{0.3\textwidth}
\vspace{-0.3cm}
\centering
\caption{Testing classes and their sizes. }  \label{tab:test_set}
\begin{tabular}{l|c}
        Scenario-Class & Size\\ 
        \hline
        1~-~$S_2$ \& $S_2$ & $60$ \\
        1~-~$S_2$ \& $S_3$ & $120$ \\
        (1, 2)~-~$P_3$ \& $S_2$ & $60$ \\ \hline
        2~-~$P_3$ & $30$  \\ \hline
        3~-~$S_3$ & $60$ \\
    \end{tabular}
\end{wraptable}

\paragraph{Transfer Scenarios. } To assert the transfer capabilities of the different agents, we investigate how \gnn-based agents with semantic goal spaces to goal configurations used as test goals though they have never trained on them before. To this end, we consider three different scenarios:
\textbf{1) Transfer to combinations of constructions}, where
agents are prevented from training on any goal including combinations of known constructions. The corresponding set of test goals includes the classes $S_2$\&$S_2$, $P_3$\&$S_2$ and $S_2$\&$S_3$; \textbf{2) Transfer to pyramids}, where
agents are prevented from training on any goal including pyramids. The underlying set of test goals includes the classes $P_3$, and $P_3$\&$S_2$;
\textbf{3) Transfer to higher stacks}, where agents are prevented from training on goals including stacks of size 3 and more. The set of corresponding test goals includes $S_3$. See Table~\ref{tab:test_set} for details.

In practice, we make sure none of the testing goals are sampled during training by simply preventing any episode where a test goal was encountered (at any time step) from getting stored in the replay buffer. This also prevents \her’s future strategy from selecting these goals during replay. The set of training goals for each scenario may include all the other possibly encountered configurations. 

\subsection{Global performance metrics}
\label{sec:global_perf}
\begin{figure}[!hbt]
  \centering
  \subfloat[\label{fig:global_semantic}]{\includegraphics[width=0.4\linewidth]{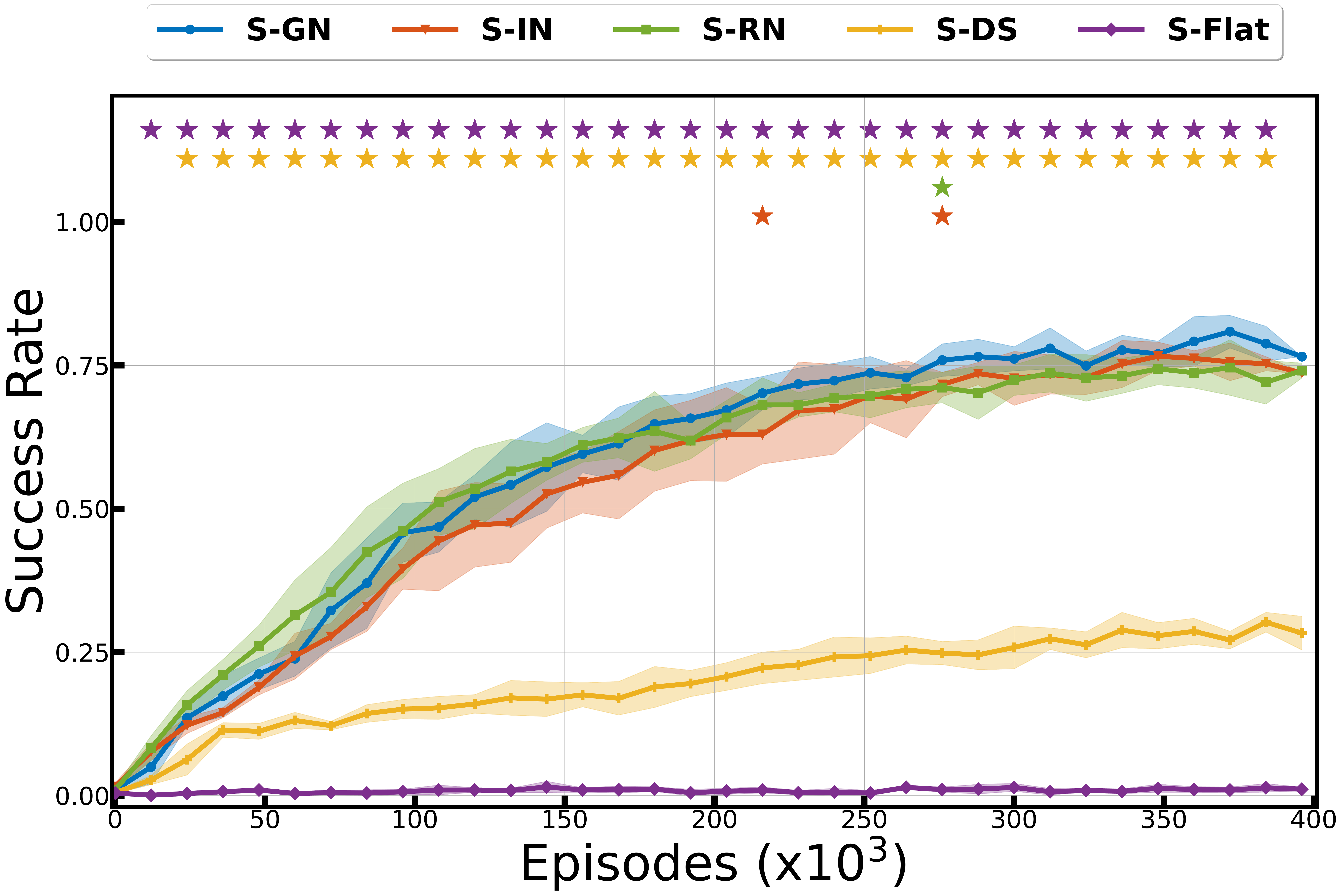}}
  \subfloat[\label{fig:global_continuous}]{\includegraphics[width=0.395\linewidth]{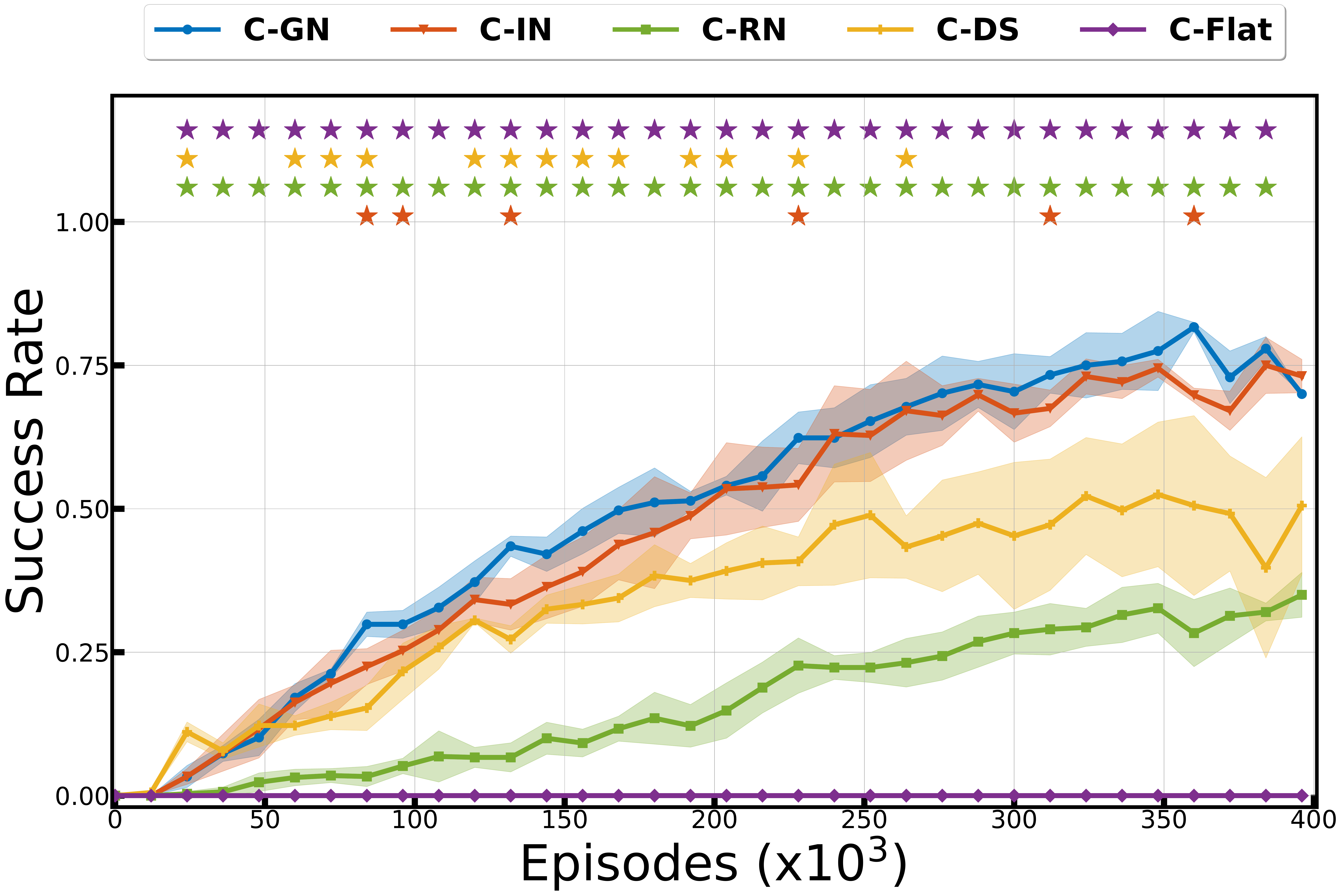}}
  \caption{Global \sr across training episodes with (a) Semantic (S) goals and (b) Continuous (C) goals for the considered agents. Mean $\pm$ standard deviations are computed over 5 seeds. Stars highlight statistical differences w.r.t \sgn agents (Welch's t-test with null hypothesis $\mathcal{H}_0$: no difference in the means, $\alpha~=~0.05$).}
  \label{fig:global_perf}
\end{figure}

In this section, we study the global performance of the different graph-based autotelic agents. Figure~\ref{fig:global_perf} presents the average \sr across evaluation classes for both semantic goals (Figure~\ref{fig:global_semantic}) and continuous goals (Figure~\ref{fig:global_continuous}). 

\paragraph{Semantic Goals. } The \sflat agents are not able to learn to reach any semantic configuration as their global \sr does not increase during training (Figure~\ref{fig:global_semantic}, purple curve). This is not surprising since simple \mlp networks that take as input high dimensional concatenations of multiple objects struggle in disentangling the learned representations. By contrast, all the \gnn-based agents are able to increase their global \sr, this is consistent with previous results showing that object-centered graph-based architectures are better suited for multi-object manipulation domains \citep{li2019towards}. On the one hand, the global \sr of \sds agents gets stuck at around 25\% (Figure~\ref{fig:global_semantic}, orange curve). This result highlights the importance of the edge update step, as it allows to focus on pairwise relations embedded within the semantic relational predicates which are used as input features to the edges. On the other hand, \sgn, \sint and \srn agents have similar performance across training episodes (Figure~\ref{fig:global_semantic}, blue, red and green curves), and show statistical differences only rarely (see stars on Figure~\ref{fig:global_semantic}). This result suggests that, when dealing with semantic relational goals, \textit{the edge update step is crucial}. In fact, \srn agents\,---\,which have a lighter computational scheme but exclusively perform pairwise edge updates\,---\,manage to perfectly catch with the more heavy architecture \sgn and \sint. 

\paragraph{Continuous Goals. }Similar to semantic goals, the \cflat agents fail to learn any interesting behavior. However, interestingly, only \gnn-based architectures that use the full computational scheme\,---\,that is, \cgn and \cint\,---\,have the best performance. On the one hand, by contrast to the semantic goals setup, \crn agents get stuck at around 30\% of the maximum global \sr (Figure~\ref{fig:global_continuous}, green curve). This suggests that the edge update step is not sufficient to capture interesting features when geometric target goals are encoded within the input edge features. On the other hand, \cds agents perform better than their \crn counterparts, with an average global \sr of 50\% (Figure~\ref{fig:global_continuous}, orange curve). This is probably due to the fact that the node update step\,---\,which is conducted by \cds but not \crn\,---\,enables information about every target geometric goal to flow to every node in the graph, which helps agents increase their performance. However, \cds agents show higher variance compared to all the other \gnn-based agents. This instability is probably explained by the role of the edge update step\,---\,which is bypassed by \cds\,---\,in disentangling useful pairwise features that help stabilize the learning. This is made even more likely as agents that perform the whole computational scheme provide the best results and the least instabilities (Figure~\ref{fig:global_continuous}, blue and red curves). 

\subsubsection{Per Class Performance Metrics}
\label{sec:local_semantic}
The global performance metrics in Section~\ref{sec:global_perf} show that the average \sr across evaluation classes gets stuck at around 75\% for both semantic and continuous agents. To investigate this, we zoom on the per class performance metrics.

\begin{figure}[!ht]
 \centering
    \includegraphics[width=0.6\textwidth]{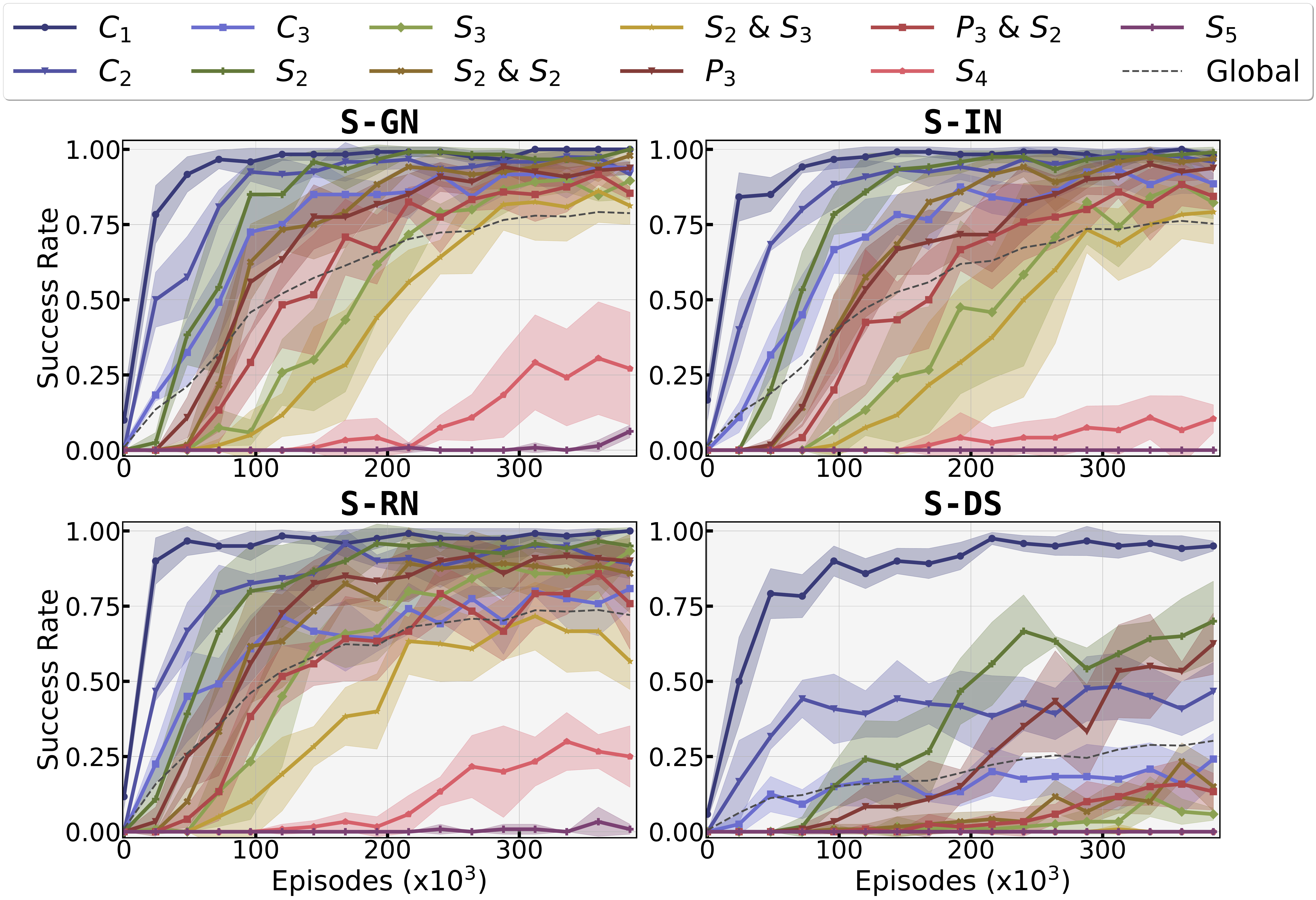}
    \caption{Local \sr for each class across training episodes with continuous goals. Mean $\pm$ standard deviations are computed over 5 seeds.}
    \label{fig:local_semantic}
\end{figure}

\begin{figure}[!ht]
 \centering
    \includegraphics[width=0.7\textwidth]{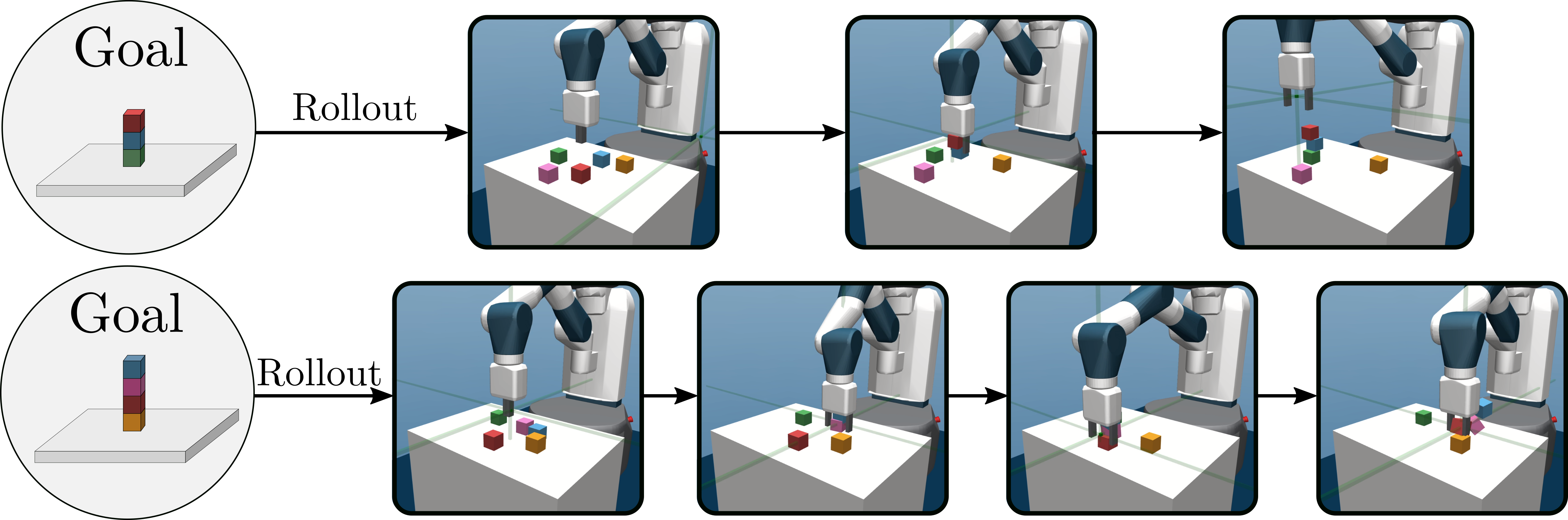}
    \caption{Example of sub-optimal behavior with semantic goals when targeting a goal in $S_3$ (up) and in $S_4$ (down). The agent tries to pick and place a stack of three objects and fails (down).}
    \label{fig:suboptimal_behavior}
\end{figure}

\paragraph{Semantic Goals. }
Figure~\ref{fig:local_semantic} shows the per-class performance of \sgn, \sint, \srn and \sds. First, and as the global performance metrics suggest (Section~\ref{sec:global_perf}), \sgn, \sint and \srn show very similar local performance. They are all able to master all the classes (with at least 65\% of \sr) except for $S_4$ and $S_5$. This failure occurs because the learned policies are sub-optimal. In fact, when rewarding themselves for each object placed correctly, the critics would most likely be \textit{greedy}: incremental rewards should come fast, even if this means not constructing stacks in the trivial order (from base upwards). As a result, agents would start by constructing the upper part of a stack, then placing it on the base object. This is not a problem for $S_3$ since robotic arms can pick and place a stack of two blocks. However, in $S_4$, it is impossible to pick and place a stack of three blocks. See Figure~\ref{fig:suboptimal_behavior} for an illustrative example. Second, the \sds agents struggle with classes that involve many constraints to be satisfied. This is because they lack enough representational power to disentangle pairwise relations between objects.

\label{sec:local_continuous}
\begin{figure}[!ht]
 \centering
    \includegraphics[width=0.6\textwidth]{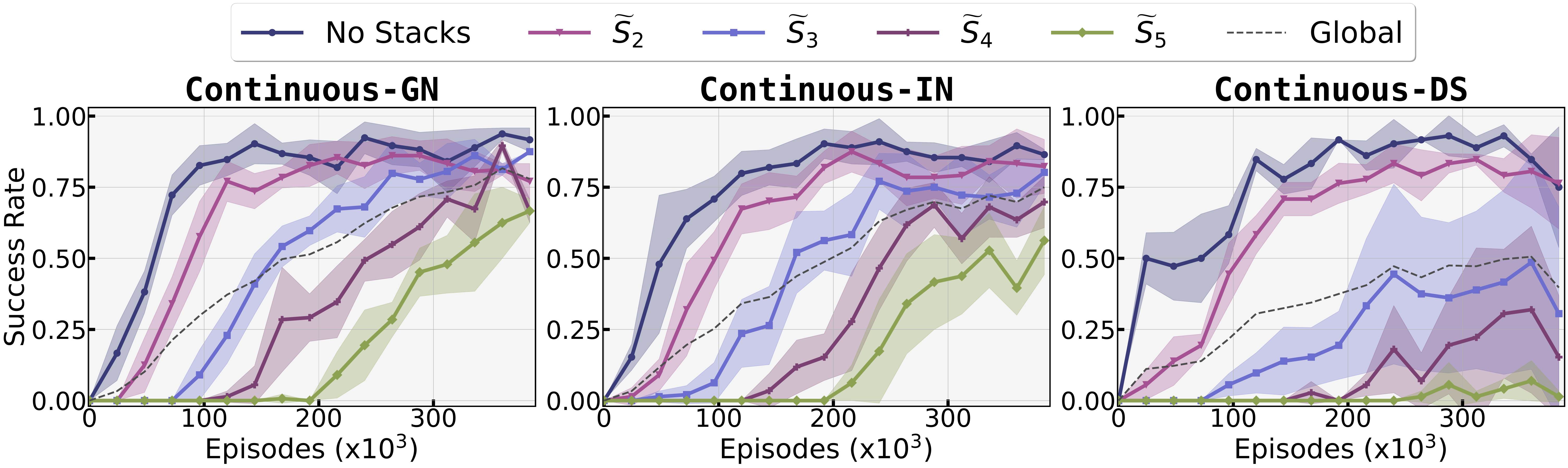}
    \caption{Local \sr for each class across training episodes with continuous goals. Mean $\pm$ standard deviations are computed over 5 seeds.}
    \label{fig:local_continuous}
\end{figure}

\paragraph{Continuous Goals. }Figure~\ref{fig:local_continuous} shows the per-class performance of \cgn, \cint and \cds. All agents first master the easy classes, before moving up to the more diffficul ones. This results from these agents leveraging automatic curriculum learning, using their \lp estimation as a proxy to choose goals that are at an affordable level of complexity. However, as opposed to semantic goals, there is less interference between classes and transfer is poorer (per-class \sr increases sequentially). On the one hand, \cds agents show a lot of instabilities beyond the $\widetilde{S}_2$ class. This further supports the idea that the node update step alone in deep sets does not provide enough representational power. On the other hand, both \cgn and \cint manage to reach goals in all the evaluation classes, from no stacks at all to stacks of 5 objects. However, they are both unable to maximize their per-class performance. This suggests that learning policies that can achieve all evaluation classes at the same time with continuous goals is difficult and requires more training budget.

\subsubsection{Transfer capabilities}
\label{sec:transfer_results}
To investigate the transfer capabilities of the \gnn-based agents, we consider the best performing architectures with semantic goal spaces: \sgn, \sint and \srn (See Section~\ref{sec:global_perf}). Tables~\ref{tab:transfer_global} and \ref{tab:transfer_local} respectively show the global and the per class performance metrics for the considered agents. The values presented in these tables correspond to the evaluation of the training policies on the held-out goals once the training is stabilized.
\begin{table}[htpb!]
\small
\vspace{-0.3cm}
    \centering
    \vspace{0.2cm}
    \begin{minipage}{0.49\linewidth}
    \caption{Global \sr metrics, averaged over $5$ seeds.\label{tab:transfer_global}}
    \resizebox{0.90\textwidth}{!}{\begin{tabular}{l|ccc}
        Agents & Scenario 1 & Scenario 2 & Scenario 3\\ 
        \hline
        \sgn & $\bm{0.93\pm0.04}$ & $0.78\pm0.09$  &  $0.28\pm0.14$  \\
        \sint & $0.89\pm0.01$ & $\bm{0.82\pm0.09}$  &  $0.47\pm0.13$ \\
        \srn & $0.89\pm0.02$ & $0.68\pm0.12$ &  $\bm{0.64\pm0.10}$ \\
    \end{tabular}}
    \end{minipage}
    \begin{minipage}{.5\linewidth}
    \caption{Per class \sr metrics, averaged over $5$ seeds.\label{tab:transfer_local}}
    \resizebox{0.95\textwidth}{!}{\begin{tabular}{l|ccc}
        Scenario-Class & \sgn & \sint & \srn\\ 
        \hline
        1~-~$S_2$ \& $S_2$ & $\bm{0.97\pm0.02}$ & $0.92\pm0.04$ &  $0.96\pm0.03$ \\
        1~-~$S_2$ \& $S_3$ & $\bm{0.89\pm0.04}$ & $0.87\pm0.01$ &  $0.82\pm0.07$ \\
        1~-~$P_3$ \& $S_2$ & $\bm{0.93\pm0.05}$ & $0.88\pm0.04$ &  $0.90\pm0.05$ \\ \hline
        2~-~$P_3$ & $0.80\pm0.09$ & $\bm{0.85\pm0.10}$ &  $0.67\pm0.14$  \\ 
        2~-~$P_3$ \& $S_2$ & $0.75\pm0.12$ & $\bm{0.80\pm0.10}$ &  $0.68\pm0.11$ \\ \hline
        3~-~$S_3$ & $0.28\pm0.14$ & $0.47\pm0.13$ &  $\bm{0.64\pm0.10}$ \\
        
    \end{tabular}}
    \end{minipage}
\end{table}

\paragraph{Transfer Scenario 1.} All the considered agents are good at transferring to combinations of constructions that they have encountered separately during training (Column 2 of Table~\ref{tab:transfer_local}). This probably results from both the representational power of \gnns and the self-attention aggregation schemes during both edge and node updates. On the one hand, all these graph-based architectures are permutation-invariant. On the other hand, for a goal within one of the combination classes, each construction is independent from the other. Consequently, agents would attend to one of them, accomplish it, and then focus on the other, as they learned to construct them separately during training.

\paragraph{Transfer Scenario 2.} During training within this scenario, agents never encounter nor train on configurations involving pyramids. In other words, an object is never placed simultaneously on top of two other objects that are close to each other. First, we compare \srn to both \sgn and \sint. \srn has lower overall transfer capabilities in this scenario compared to \sgn and \srn (Column 1 of Table~\ref{tab:transfer_global}). They struggle in transferring to held-out goals from both $P_3$ and $P_3$\&$S_2$ classes (Table~\ref{tab:transfer_local}). On the one hand, \srn agents bypass the node update step which aggregates, for a particular node, the flowing information from the other nodes. Consequently, they tend to focus more on pairwise relations and less on the global relations of a particular configuration. On the other hand, a pyramid involves three objects. In fact, to be able to put an object $i$ on top of $j$ and $k$, the latter two need to be close to each other. As a result, exclusive edge updates without allowing information to flow between the different nodes is insufficient. Second, we compare \sgn to \sint. As the former concatenates the global features to the input features of the edge update, the size of its corresponding network would be greater. This would probably lead to overfitting on the training data and transferring less to new situations.

\paragraph{Transfer Scenario 3.} During training, agents never encountered a configuration involving a stack of 3 objects. Column 3 of Table~\ref{tab:transfer_global} shows that \srn outperforms both \sgn and \sint in this scenario. This can be explained in a threefold fashion. First, we defined the \textit{above} predicate as being \textit{directly above}. Hence, a goal from the class $S_3$ would involve only two above predicates of the form \textit{above}($i$, $j$) and \textit{above}($j$, $k$)\,---\,which are fed to two different edges\,---\,to be true. Second, \srn agents focus on pairwise relations that exist within edges. In other words, if they have learned to stack two blocks, they are capable to sequentially perform two independent stacks of two. Third, a stack of three objects can be constructed independently from the order of the stacks (i.e. first $B$ above $C$ then $A$ above $B$ or first $A$ above $B$ then $A/B$ on $C$). By contrast \sgn and \sint tend to transfer less as they overfit more on the training data due to their heavier computational scheme. 


\section{Conclusion and Future Work}
In this paper, we studied several \gnn-based goal-conditioned architectures for both the policy and critic in multi-object manipulation domains. More specifically, we considered four different computational schemes: full graph networks, interaction networks, relation networks and deep sets. We evaluated our agents using two different goal space structures: 1) continuous geometric goal spaces corresponding to per-object target positions; 2) semantic relational goal spaces based on the binary predicates close and above. Finally, we studied the transfer capabilities in three different scenarios by assessing the agents performance on different sets of held-out semantic goals.

\paragraph{Semantic and Continuous Goal Spaces. } Autotelic agents benefit from the abstract structured representation within semantic goal spaces. In fact, our results show that, when dealing with continuous goals, full graph and interaction networks, which adapt the full graph computational scheme, are the best performing agents. For this particular type of goal space, information about all the objects needs to flow to each node in order to learn more complex goals. By contrast, with semantic goal spaces, performing the edge update step is sufficient to capture useful pair-wise relations between objects. In fact, relation networks perform in par with full graph and interaction networks. Finally, additional ingredients such as non-trivial scene resets and \lp-based goal selection were necessary for agents to learn complex continuous goals. However, their counterparts that use semantic goals are more flexible as they do not need these additional ingredients. 

\paragraph{Transfer Capabilities with Semantic Goals. } Our transfer study suggests three main results. First, full graph, interaction and relation networks are all able to transfer to combinations of previously seen goals. Second, relation networks struggle in adapting to new goals that require reasoning about triplets of objects, such as building pyramids. These agents bypass the node update step, which seems to be crucial to pool information about all the nodes in the graph. Finally, relation networks outperform the other \gnn-based architecture in transferring to goals of higher stacks. In fact, their light computational scheme enables them to overfit less on the training data. Consequently, they are more flexible in sequentially combining pair-wise skills. 

This study helps understand the impact of key design choices towards open-ended autotelic agents capable of efficient transfer between abstract goals. However, the agents studied here only leverage their physical interactions with the environment. This does not account for the extraordinary human capacities to learn from social interactions \citep{Vygotskii1978, bruner1973organization, tomasello2009constructing}.
We believe adding social learning mechanisms as suggested in  \citep{sigaud2021towards} is a promising line of research towards more capable open-ended agents. 

\subsubsection*{Acknowledgments}
This work was performed using HPC resources from GENCI-IDRIS (Grant 2020-A0091011875).

\bibliography{collas2022_conference}
\bibliographystyle{collas2022_conference}

\appendix
\section{Appendix}

\subsection{Implementation details}
\label{supp:implementation_details}
In this part, we present details necessary to reproduce our results. We further open-source our code at \href{https://github.com/akakzia/rlgraph_2}{https://github.com/akakzia/rlgraph\_2}. 

\textit{\gnn-based networks.} Our four graph-based architectures use at most two shared networks, $NN_\text{edge}$ and $NN_\text{node}$, respectively for computing updated edge features and node features. Both are 1-hidden-layer networks of hidden size $256$. Taking the output dimension to be equal to 3$\times$ the input dimension for the shared networks showed better results. All networks use ReLU activation and the Xavier initialization. For edge-wise and node-wise aggregation, we use a one-headed self-attention module. Finally, to produce the output, all architecture use a readout network $NN_\text{out}$. The latter is also a 1-hidden-layer network of hidden size $256$. For optimization, we use Adam with learning rates $10^{-3}$. The list of hyperparameters is provided in Table~\ref{tab:hyperparams}.

\textit{Parallel implementation of \sac-\her.} All our experiments are based on a \emph{Message Passing Interface} \citep{dalcin2011parallel} to exploit multiple processors. Each of the $24$ parallel workers maintains its own replay buffer of size $10^6$ and performs its own updates. To synchronize experience between different workers, updates are summed over the $24$ actors and the updated actor and critic networks are broadcast to all workers. Each worker alternates between $2$ data collection episodes and $30$ updates with batch size $256$. To form an epoch, this cycle is repeated $50$ times and followed by the offline evaluation of the agent. 
\begin{table}[htbp!]
    \centering
    \caption{Hyperparameters used in this paper.\label{tab:hyperparams}}
    \vspace{0.2cm}
    \begin{tabular}{l|c|c}
        Hyperparam. &  Description & Values. \\
        \hline
        $nb\_mpis$ & Number of workers & $24$ \\
        $nb\_cycles$ & Number of repeated cycles per epoch & $50$ \\
        $nb\_rollouts\_per\_mpi$ & Number of rollouts per worker & $2$ \\
        $rollouts\_length$ & Number of episode steps per rollout & $200$ \\
        $nb\_updates$ & Number of updates per cycle & $30$ \\
        $replay\_strategy$ & \her replay strategy & $future$ \\
        $k\_replay$ & Ratio of \her data to data from normal experience & $4$ \\
        $batch\_size$ & Size of the batch during updates & $256$ \\
        $\gamma$ & Discount factor to model uncertainty about future decisions & $0.99$ \\
        $\tau$ & Polyak coefficient for target critics smoothing & $0.95$ \\
        $lr\_actor$ & Actor learning rate & $10^{-3}$ \\
        $lr\_critic$ & Critic learning rate & $10^{-3}$ \\
        $\alpha$ & Entropy coefficient used in \sac  & $0.2$ \\
        $biased\_init$ & Probability of following non-trivial scene reset scheme & $0.2$ \\
        $self\_eval\_curriculum$ & Probability to perform self evaluations & $0.1$ \\
        $curriculum\_queue\_length$ & Window over which \lp estimations are made & $1000$ \\
    \end{tabular}
\end{table}

\section{Additional results}
\label{supp:additional}
In this section, we present additional results which complement the ones presented in the main paper. More specifically, we study the relative importance of curriculum learning when using continuous goals (Appendix~\ref{supp:add_curriculum}) and of self-attention when using semantic goals (Appendix~\ref{supp:add_attention}). 

\subsection{Curriculum ablation}
\label{supp:add_curriculum}
\begin{figure}[!hbt]
  \centering
  \subfloat[Global \sr. Stars highlight statistical differences w.r.t \cgn agents. \label{fig:global_zpd_abla}]{\includegraphics[width=0.44\linewidth]{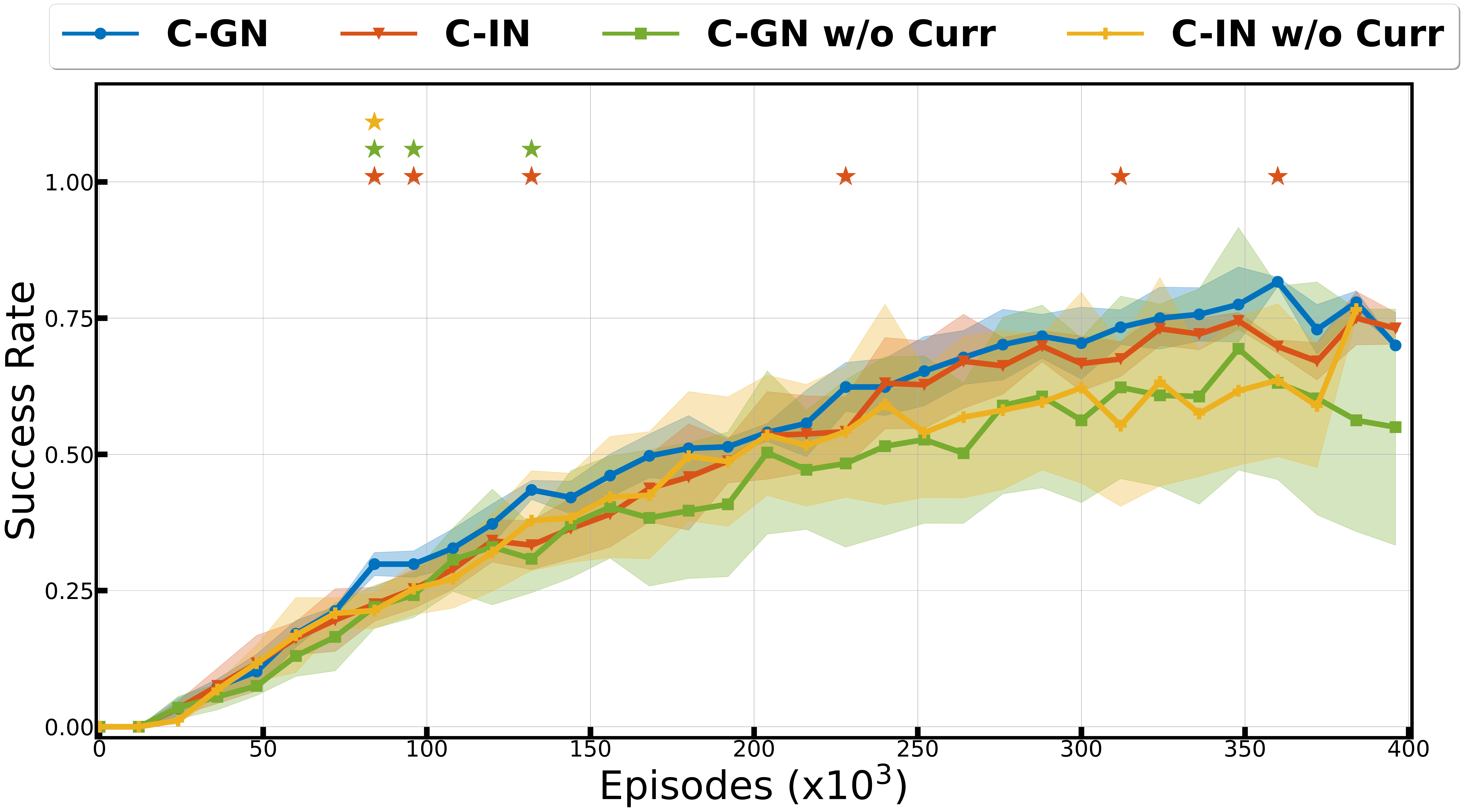}}
  \subfloat[Per class \sr. \label{fig:local_zpd_abla}]{\includegraphics[width=0.5\linewidth]{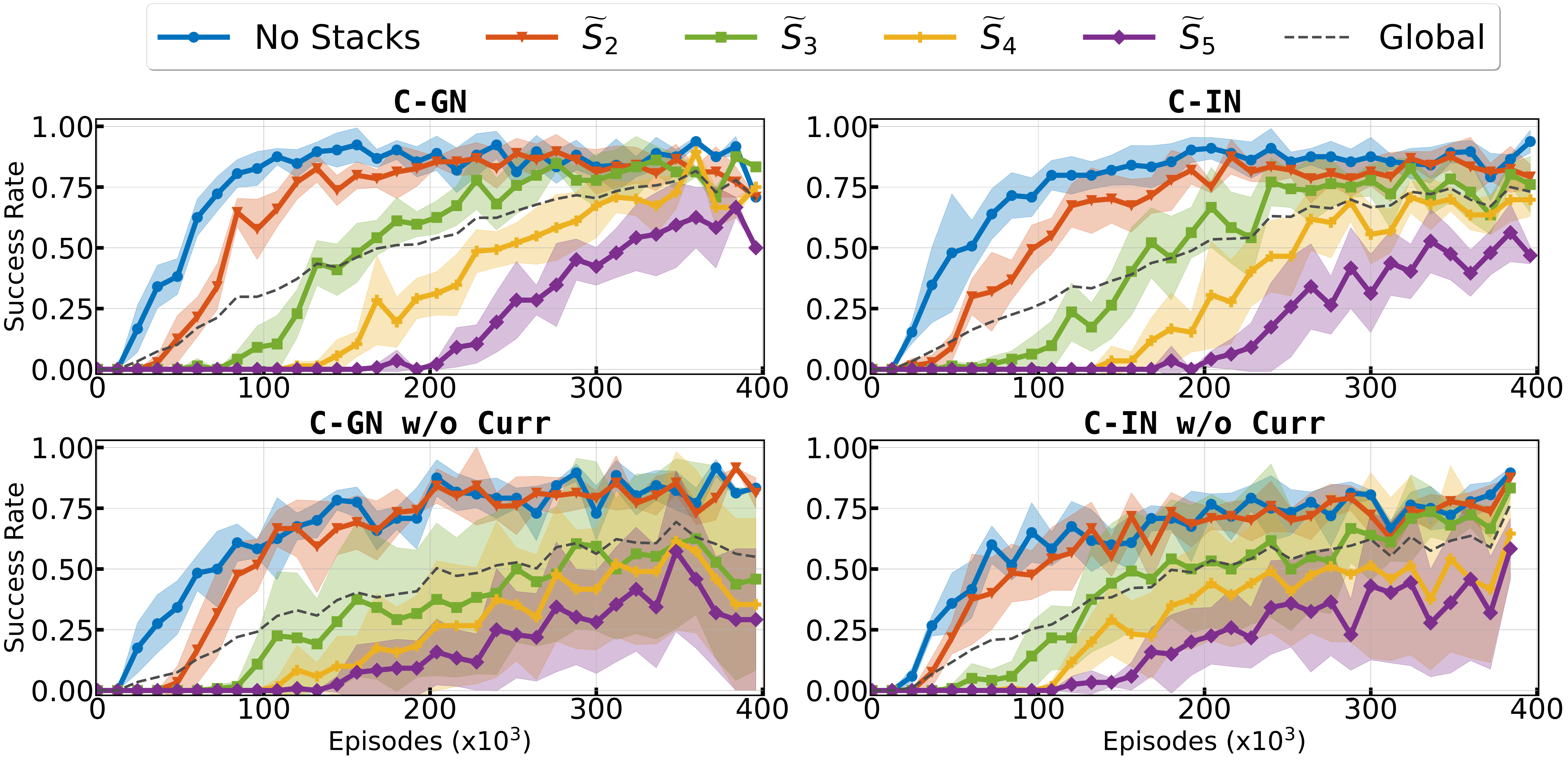}}
  \caption{Performance metrics for \cgn, \cint and their curriculum ablations. Mean $\pm$ standard deviations are computed over 5 seeds.}
  \label{fig:abla}
\end{figure}

To study the relative importance of the \lp-based curriculum learning mechanism used with continuous goals, we introduce ablations of \cgn and \cint which uniformly sample a class of goals without any particular prioritization. We only consider architectures based on \gn and \inter in this ablation study since they show the best results in Section~\ref{sec:global_perf}. Figure~\ref{fig:abla} presents the global performance metrics for \cgn, \cint and their ablation counterparts. Autotelic agents using continuous goals but no curriculum clearly show an increased variance in their global performance. Figure~\ref{fig:abla} zooms on the local performance on each class for the considered agents. Compared to \cgn and \cint, the shaded areas in the ablations are larger, suggesting that the learning process of the latter agents is not stable. Precisely, this is true in stacks of size 3 or higher. In fact, ablations face catastrophic forgetting as they engage with harder goals. The curriculum learning mechanism helps stabilize the learning process by focusing on goals of moderate level of complexity, including the ones that the agents are likely to forget during training. Note that this issue is specific to continuous goals, which shows that they are not well suited to transfer between different goals.

\subsection{Self-attention ablation}
\label{supp:add_attention}
\begin{figure}[!ht]
 \centering
    \includegraphics[width=0.5\textwidth]{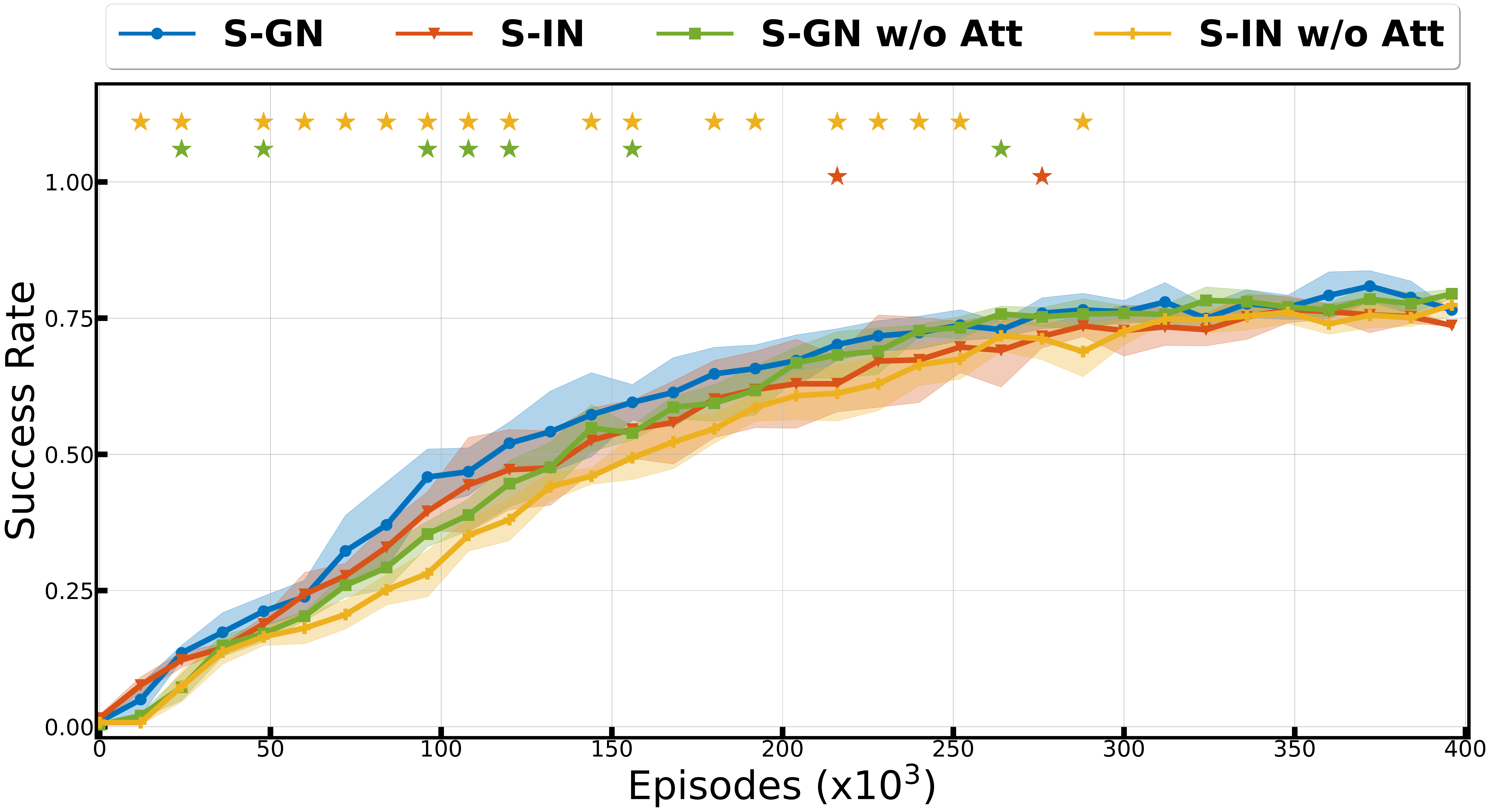}
    \caption{Global \sr across training episodes for \sgn, \sint and their self-attention ablations counterparts. Mean $\pm$ standard deviations are computed over 5 seeds. Stars highlight statistical differences w.r.t \sgn agents (Welch's t-test with null hypothesis $\mathcal{H}_0$: no difference in the means, $\alpha~=~0.05$.}
    \label{fig:ablation_attention}
\end{figure}

We propose to remove the self-attention aggregation schemes from \sgn and \sint,\---\,the two best performing agents,\---\,and introduce the corresponding ablations which use an unweighted sum when performing the pooling over edges or nodes. Figure~\ref{fig:ablation_attention} presents the global \sr for these agents across training episodes. The differences only appear at the beginning of training. In fact, the global performance metrics in the ablations increases slower than their corresponding full-versions (blue vs green; red vs orange). However, all agents seem to behave similarly by the end of training. This suggests that self-attention improves sample efficiency, yielding \gnn-based agents that can faster capture actionable relational features within their graphs. 

\subsection{Non-trivial scene reset ablation}
\label{supp:zpd_ablation}
\begin{figure}[!hbt]
  \centering
  \subfloat[Global \sr. Stars highlight statistical differences w.r.t \cgn agents. \label{fig:global_zpd}]{\includegraphics[width=0.44\linewidth]{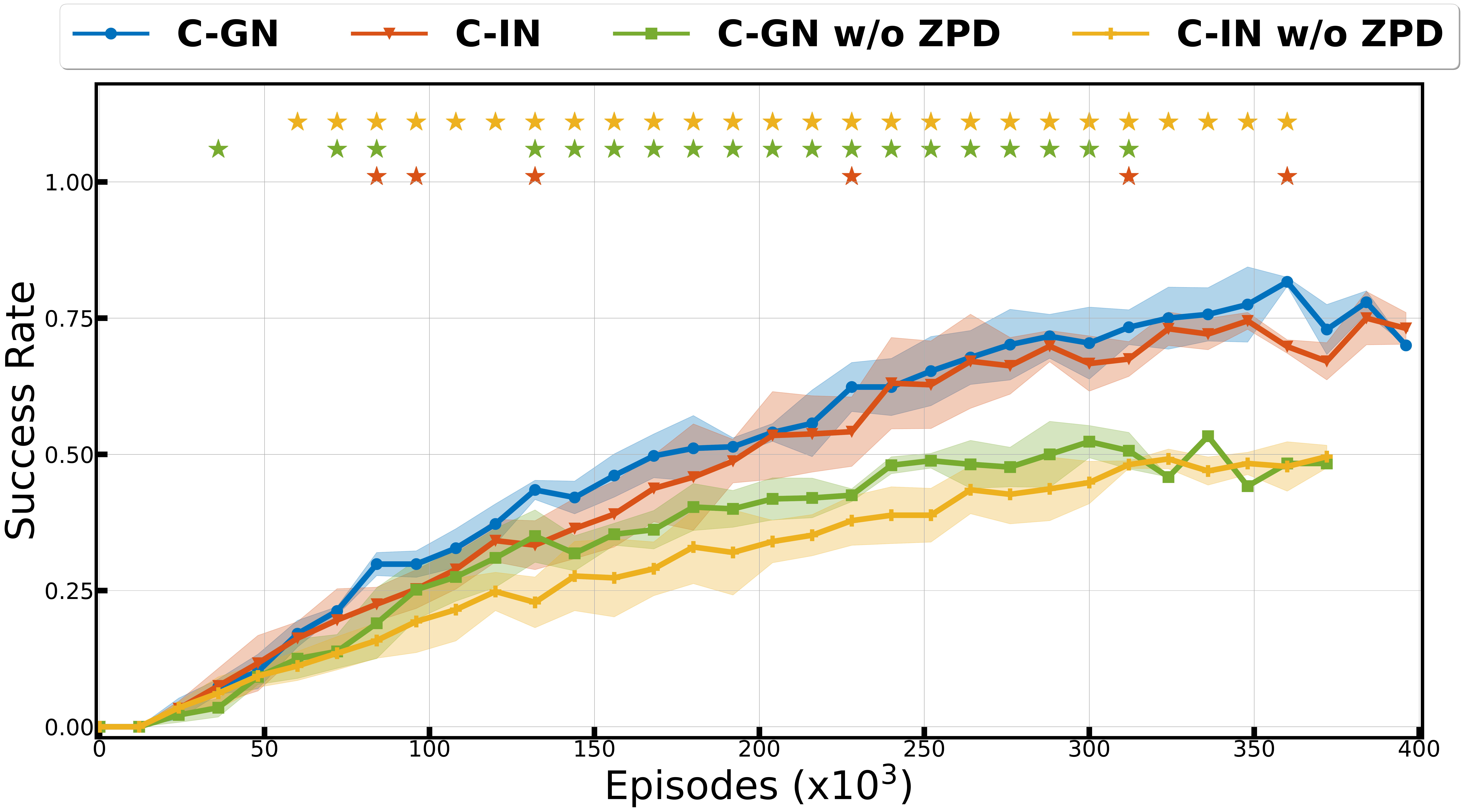}}
  \subfloat[Per class \sr. \label{fig:local_zpd}]{\includegraphics[width=0.5\linewidth]{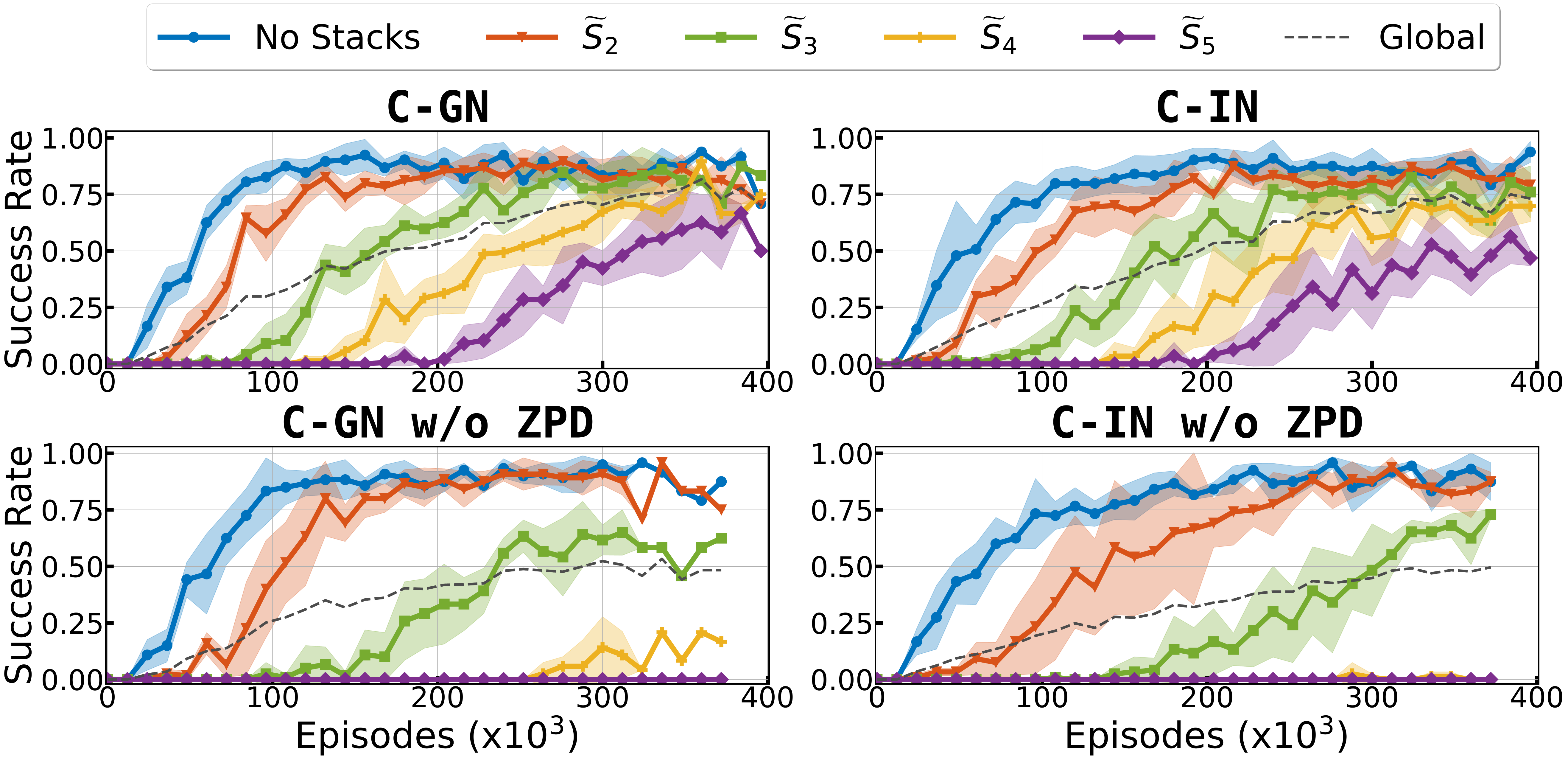}}
  \caption{Performance metrics for \cgn, \cint and their ablations where non-trivial scene reset is removed. Mean $\pm$ standard deviations are computed over 5 seeds.}
  \label{fig:main_ab_zpd}
\end{figure}

To assess the importance of the non-trivial scene reset scheme for continuous goals, we consider the \cgn and \cint agents\,---\,the best performing \gnn-based architectures so far\,---\, and remove the biased initialization scheme: blocks are placed without any initial stacks in the resulting ablations. Figure~\ref{fig:main_ab_zpd} shows performance metrics for these agents. The global \sr of both ablations increases slower than that of \cgn and \cint (\figurename~\ref{fig:local_zpd}). Besides, it gets stuck at around 50\% of the maximal performance while their corresponding full versions manage to reach 75\%. Zooming on the per-class performance metrics shows the considerable decrease in the capability to reach complex goals when removing the non-trivial reset scheme (\figurename~\ref{fig:local_zpd}): the ablations struggle to transfer between easy goals ($\widetilde{S}_2$ and $\widetilde{S}_3$) and harder ones ($\widetilde{S}_4$ and $\widetilde{S}_5$).
\end{document}